%% file: ECCV_2022.tex
\documentclass[runningheads]{ECCV/llncs}
\usepackage{graphicx}

\usepackage{tikz}
\usepackage{comment}
\usepackage{amsmath,amssymb} 
\usepackage{color, colortbl}

\usepackage{blindtext}
\usepackage[caption=false]{subfig}
\usepackage{multirow}
\usepackage{url}

\definecolor{darkseagreen}{rgb}{0.56, 0.74, 0.56}
\definecolor{grannysmithapple}{rgb}{0.66, 0.89, 0.63}
\definecolor{aliceblue}{rgb}{0.94, 0.97, 1.0}

\usepackage[accsupp]{axessibility}  

\begin{document}
\pagestyle{headings}
\mainmatter
\def\ECCVSubNumber{1455}  

\title{Gradient-based Uncertainty for Monocular Depth Estimation} 

\titlerunning{Gradient-based Uncertainty for Monocular Depth Estimation}
%
\author{Julia Hornauer\inst{1} \and
Vasileios Belagiannis\inst{2}\thanks{Most of this work was done while Vasileios Belagiannis was with Ulm
University.}}
\authorrunning{J. Hornauer et al.}
%
\institute{Institute of Measurement, Control and Microtechnology, Ulm University, Germany \\
\email{julia.hornauer@uni-ulm.de}
\and Department of Simulation and Graphics, Otto von Guericke University
Magdeburg, Germany}
\maketitle

\begin{abstract}
In monocular depth estimation, disturbances in the image context, like moving objects or reflecting materials, can easily lead to erroneous predictions. For that reason, uncertainty estimates for each pixel are necessary, in particular for safety-critical applications such as automated driving. We propose a post hoc uncertainty estimation approach for an already trained and thus fixed depth estimation model, represented by a deep neural network. The uncertainty is estimated with the gradients which are extracted with an auxiliary loss function. To avoid relying on ground-truth information for the loss definition, we present an auxiliary loss function based on the correspondence of the depth prediction for an image and its horizontally flipped counterpart. Our approach achieves state-of-the-art uncertainty estimation results on the KITTI and NYU Depth V2 benchmarks without the need to retrain the neural network. Models and code are publicly available at \url{https://github.com/jhornauer/GrUMoDepth}.
\keywords{Depth Estimation, Uncertainty Estimation, Training-Free}
\end{abstract}

\input{Sections/introduction}
\input{Sections/related_work}
\input{Sections/method}
\input{Sections/experiments}

\section{Conclusions}
We present the usage of gradients for uncertainty estimation of depth predictions. Given an already trained model, we introduce an auxiliary loss function, that is independent of ground truth depth, based on the correspondence of the depth prediction for an image and its horizontally flipped counterpart. Note that our method does not adjust the neural network weights but estimates the uncertainty in a \textit{post hoc} manner. With the loss defined as the error between the depth predictions, we calculate the derivative \textit{w.r.t.} feature maps by backpropagation. Finally, we determine the pixel-wise uncertainty based on the gradients of an inter-mediate feature map extracted from the depth decoder. We performed a comprehensive comparison with related work using two depth estimation benchmarks, namely the KITTI and NYU Depth V2 datasets. Our evaluation shows the high potential to estimate uncertainty based on gradients extracted from an already trained model. 

\paragraph{\textbf{Acknowledgements.}} The research leading to these results is funded by the German Federal Ministry for Economic Affairs and Climate Action" within the project “KI Delta Learning“ (Förderkennzeichen 19A19013A). The authors would like to thank the consortium for the successful cooperation.

%
%
\bibliographystyle{ECCV/splncs04}
\bibliography{bibfile}
\end{document}

%% file: Sections/introduction.tex
\section{Introduction}
Deep neural networks have shown astonishing performance in 3D perception tasks such as depth prediction~\cite{monodepth2,Laina2016DeeperDP}. Depth estimation from a single image has particularly attracted attention because RGB cameras are cheaper compared to LiDAR sensors while offering higher resolution and frame rates. Nevertheless, disturbances in the image context, like occlusions, moving objects, or reflecting materials, can easily affect the neural network and therefore lead to erroneous predictions~\cite{Nguyen_2015_CVPR}. It is thus crucial to estimate the uncertainty of the depth estimates, especially for safety-critical applications such as automated driving.

Bootstrapped ensembles~\cite{Lakshminarayanan2017SimpleAS} and Monte Carlo Dropout~\cite{mcdropout}, which are the well-known uncertainty estimation approaches, have been explored for computationally expensive tasks, such as depth estimation~\cite{Poggi2020OnTU}. 
Both approaches rely on sampling from the model distribution, which is not suitable for real-time applications due to the high computational cost.
Predictive methods, such as maximum likelihood maximization~\cite{Kendall2017WhatUD,Klodt2018SupervisingTN}, on the other hand, require the adaptation of the training procedure to learn an estimate of the uncertainty contained in the data. Also, it is not always wished or even possible to retrain the neural network with the objective of uncertainty estimation. For instance, the parameters of models provided by external sources are usually not accessible and therefore not modifiable. The parameter modification is not feasible for neural network models that are either specialized to the target system (e.g., by pruning) or have to meet certain system requirements. For these reasons, we target the problem of uncertainty estimation of an already trained model in a \textit{post hoc} manner.

One approach to training-free uncertainty estimation is the usage of dropout only during inference~\cite{Mi2019TrainingFreeUE}. Although this approach does not require any adaptation of the training protocol, the uncertainty estimation is still based on sampling from the model distribution and is therefore not suitable for real-time applications. In contrast, we propose to estimate uncertainty based on gradients extracted from the neural network, which is much less computationally intensive and requires only one additional backward pass. 

Inspired by gradient-based approaches for out-of-distribution detection in classification~\cite{Lee2020GradientsAA,Huang2021OnTI}, we use gradients to estimate the uncertainty associated with pixel-wise depth, obtained from monocular depth estimation models. Given a trained model, we aim to extract meaningful gradients without relying on the ground-truth depth values. Therefore, we define an auxiliary loss function based on the correspondence of the depth prediction for an image and its horizontally flipped counterpart. With the loss defined as the error between both depth predictions, we calculate the derivative \textit{w.r.t.} feature maps, by back-propagating through the neural network. 
We then rely on the gradients extracted from the feature maps of a decoder layer of the depth estimation model to obtain the final uncertainty score. 
Importantly, our approach achieves state-of-the-art uncertainty estimation results without the need to retrain the neural network.  

Overall, we summarize the contributions of our paper as follows: 
Firstly, we propose a \textit{post hoc} uncertainty estimation approach for depth prediction based on the gradients extracted from already trained models. Essentially, our approach is independent of how the model was trained.
Secondly, for the gradient generation, we define an auxiliary loss function without relying on ground-truth depth values. In this context, we empirically show the meaningfulness of the gradient-based uncertainty estimation.
Lastly, in an extensive comparison to existing approaches on the two common depth estimation benchmarks KITTI~\cite{Geiger2013IJRR} and NYU Depth V2~\cite{SilbermanECCV12}, we demonstrate \textit{state-of-the-art} uncertainty estimation results. 

%% file: Sections/related_work.tex
\section{Related Work}
\paragraph{\textbf{Depth Estimation Uncertainty}}
Compared to classification tasks, it is more difficult to determine the uncertainty of high-dimensional predictions, such as in depth estimation. 
In general, a distinction can be made between epistemic and aleatoric uncertainty. While epistemic uncertainty arises from model weights and is reducible with more training data, aleatoric uncertainty is due to noise in the input data \cite{Kendall2017WhatUD}.
Bootstrapped ensembles~\cite{Lakshminarayanan2017SimpleAS} and Monte Carlo (MC) Dropout~\cite{mcdropout} estimate the epistemic uncertainty by modeling the distribution over parameters. For bootstrapped ensembles, this is achieved by training multiple models, sampling initial weights from a specified distribution, while for MC Dropout dropout layers are applied during training and inference. 
The depth and uncertainty estimates are obtained by sampling from the model distribution and computing the mean and variance, respectively.
A predictive approach that accounts for aleatory uncertainty is to learn a distribution with mean and variance that represents the data-dependent error instead of a single output value by maximizing the negative log-likelihood~\cite{Klodt2018SupervisingTN}. 
Kendall and Gal~\cite{Kendall2017WhatUD} demonstrate how to combine both types of uncertainty. 
Recent works explore the integration of uncertainty for computationally intensive tasks such as depth estimation~\cite{Chanduri2021CamLessMonoDepthMD,Yang2020D3VODD}, semantic segmentation~\cite{Gustafsson2020EvaluatingSB}, optical flow~\cite{Wannenwetsch2017ProbFlowJO}, or multi-task learning~\cite{Zhou2021SUBDepthSA}.
Moreover, Poggi et al.~\cite{Poggi2020OnTU} extensively compare different empirical and predictive uncertainty estimation approaches for self-supervised depth estimation.
Among others, image flipping post-processing proposed by Godard et al.~\cite{Godard2017UnsupervisedMD} serves as a simple baseline to obtain the variance over two outputs as an uncertainty measure. 
In addition, Poggi et al.~\cite{Poggi2020OnTU} present their approach of self-training, where variance learned with log-likelihood maximization is improved by using knowledge obtained from a teacher model using its predictions as labels.
One downside of those approaches is the adaptation of the model design~\cite{Klodt2018SupervisingTN} or the specific training pipeline~\cite{mcdropout,Lakshminarayanan2017SimpleAS}. Furthermore, empirical approaches have additional computational overhead~\cite{mcdropout,Lakshminarayanan2017SimpleAS} and increased memory footprint~\cite{Lakshminarayanan2017SimpleAS}, making them unsuitable for real-time applications such as autonomous driving or robotics. By contrast, in this work, we explore the usage of gradients as a \textit{post hoc} uncertainty estimation approach that is independent of the conducted training procedure. 
Since model re-training is not always feasible, Mi et al.~\cite{Mi2019TrainingFreeUE} explore different training-free strategies to generate a distribution over the model output by data augmentation, inference-time dropout, and additive noise in intermediate network layers. 
We also explore training-free uncertainty estimation but forgo computationally intensive sampling by relying on gradients extracted with an auxiliary loss function. 

\paragraph{\textbf{Model Robustness by Gradient Analysis}}
Deep neural networks are mostly trained using gradient-based optimization.
The informativeness of gradients for the task of out-of-distribution detection is explored in~\cite{Oberdiek2018ClassificationUO,Lee2020GradientsAA,Huang2021OnTI}.
To obtain the gradients, a loss function must be defined before backpropagating through the model. Oberdiek et al.~\cite{Oberdiek2018ClassificationUO} use the negative log-likelihood at the predicted class since no labels are present during testing.
Lee and AlRegib~\cite{Lee2020GradientsAA}, on the other hand, define confounding labels with only positive entries for the cross-entropy loss to show whether the model can associate features with any of the learned classes. Huang et al.~\cite{Huang2021OnTI} determine the KL divergence between the softmax output and the Uniform distribution, which indicates whether the predicted probabilities are distributed across all classes or concentrated in one class. Unlike, we use gradients of a regression neural network to determine the pixel-wise uncertainty for depth estimation models. More precisely, we need an uncertainty score for each input pixel and not just one score for the entire input image. For this purpose, we define the pixel-wise distance between the depth prediction and a reference depth obtained by image transformation as an auxiliary loss function.

\paragraph{\textbf{Monocular Depth Estimation}}
The recent monocular depth estimation approaches train deep neural networks with supervision~\cite{Eigen2014DepthMP,Laina2016DeeperDP,Bauer2021NVSMonoDepthIM,Bhat2021AdaBinsDE,Song2021MonocularDE}. While Eigen et al.~\cite{Eigen2014DepthMP} use information from local and global features, Laina et al.~\cite{Laina2016DeeperDP} propose a fully convolutional deep neural network and Bauer et al.~\cite{Bauer2021NVSMonoDepthIM} integrate novel view synthesis. 
However, a large amount of data, which is time-consuming and expensive to obtain, is necessary for neural network training. Especially the generation of ground truth depth requires different well calibrated sensors. Therefore, self-supervised approaches leveraging monocular sequences~\cite{monodepth2,Zhou2017UnsupervisedLO}, or stereo image pairs~\cite{Godard2017UnsupervisedMD,Yang2018DeepVS,zhou_diffnet} are proposed. In the works of Godard et al.~\cite{Godard2017UnsupervisedMD} as well as Yang et al.~\cite{Yang2018DeepVS}, learning depth from stereo image pairs by leveraging the scene geometry is introduced. 
In contrast, the use of monocular sequences for supervision by image reprojection~\cite{Zhou2017UnsupervisedLO}, which is beneficial in practical applications, is more challenging due to the unknown scale and camera position that must be learned simultaneously with depth.
Godard et al.~\cite{monodepth2} adapt the reprojection loss to handle occlusions and moving objects, while semantic or scale consistency is proposed by Toasi et al.~\cite{Tosi2019LearningMD} and Bian et al.~\cite{Bian2019UnsupervisedSD}, respectively.
Xu et al.~\cite{Xu2019RegionDN}, on the other hand, introduce Region Deformer Networks to take moving objects into account. We propose an uncertainty estimation approach for monocular depth estimation models that are trained in a supervised or self-supervised manner. For the self-supervised case, the supervision can be provided by both monocular as well as stereo image pairs. 

%% file: Sections/method.tex
\section{Method}
Consider a deep neural network $\mathbf{d} = f(\mathbf{x}; \theta)$, parameterized by $\theta$, that takes an image $\mathbf{x} \in \mathbb{R}^{w \times h \times 3}$ with width $w$ and height $h$ as input and predicts the pixel-wise depth $\mathbf{d} \in \mathbb{R}^{w \times h \times 1}$. Based on the trained depth prediction model $f_{\theta}$, we aim to predict the uncertainty $\mathbf{u} \in \mathbb{R}^{w \times h \times 1}$ for each predicted depth value.

\begin{figure}[ht]
    \centering
    \includegraphics[width=\textwidth]{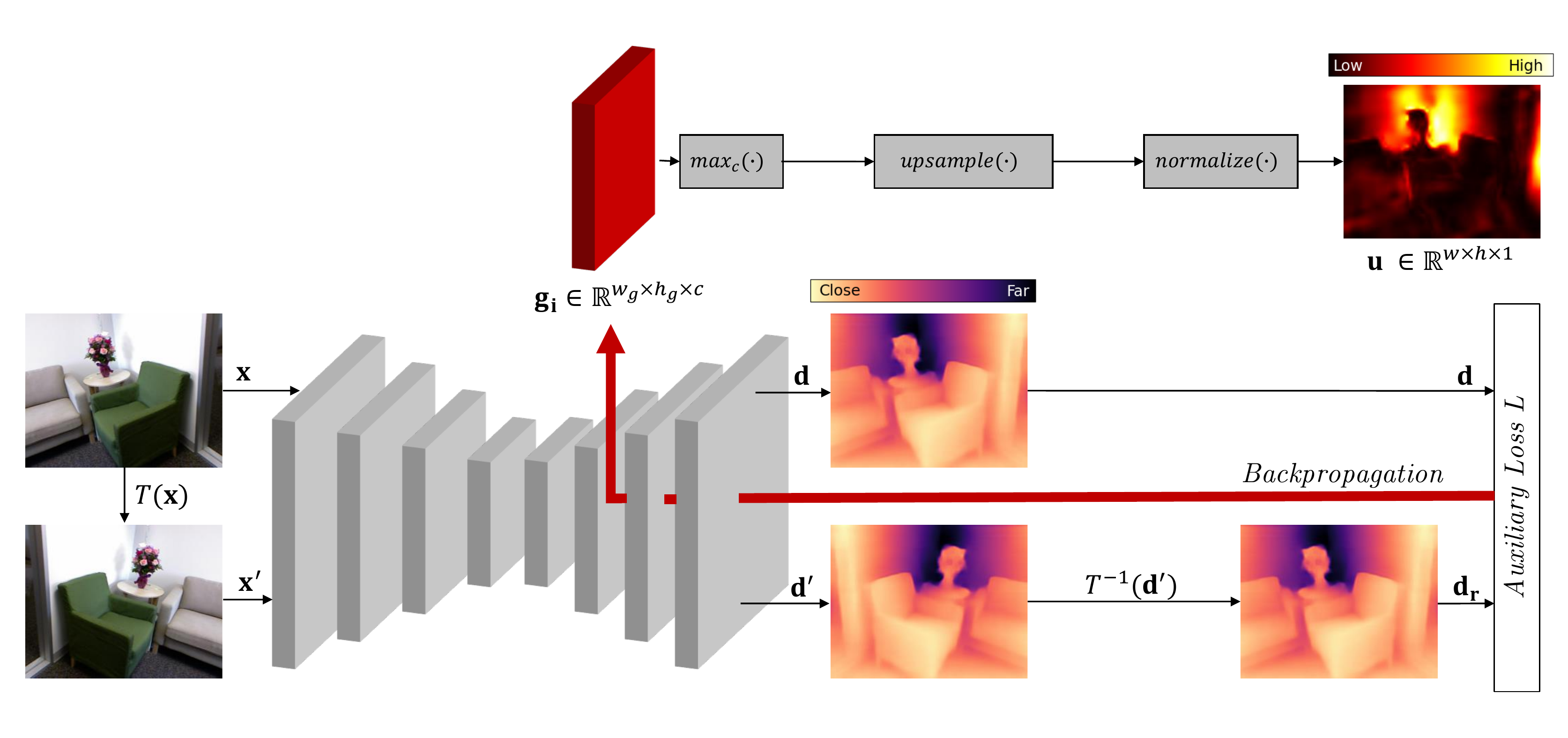}
    \caption{First, the image $\mathbf{x}$ and its flipped counterpart $\mathbf{x}^{\prime}$ are forwarded through the neural network. Then, the depth prediction $\mathbf{d}^{\prime}$ of the flipped counterpart is flipped back to obtain the reference depth $\mathbf{d}_{r}$. Afterwards an auxiliary loss based on the two depth predictions is calculated to back-propagate through the neural network. Finally, the gradients $\mathbf{g}_{i}$ of the $i$-th intermediate layer are extracted. To determine the pixel-wise uncertainty score $\mathbf{u}$, the channel number is reduced from $c$ to $1$ by taking the maximum value of each pixel in $\mathbf{g}_{i}$ in the channel dimension. Then, the gradient map is upsampled with nearest neighbor interpolation to match the depth map resolution.}
    \label{fig:overview}
\end{figure}

\paragraph{\textbf{Post Hoc Uncertainty Estimation}}
In this work, we target \textit{post hoc} uncertainty estimation. We do not adjust the model parameters, but only assume access to the internal model representation given by the feature maps $\mathbf{a}_{i}$, where $i$ denotes the feature map obtained from the $i$-th layer of the depth decoder. Essentially, we estimate the uncertainty of the already trained depth estimation model and therefore do not need to adjust the training protocol or the network design. Moreover, our approach is independent of the underlying model training strategy. For instance, the depth estimation model may have been trained in a supervised or self-supervised manner with monocular or stereo supervision. Fig.~\ref{fig:overview} illustrates an overview of our method. 

\subsection{Gradient-based Uncertainty}
\paragraph{\textbf{Gradient Generation}}
We propose to use gradients extracted from the feature maps $\mathbf{a}_{i}$ to estimate the pixel-wise uncertainty of depth predictions. For the gradient generation, the pixel-level loss function must be defined first, followed by the derivative of the loss function \textit{w.r.t} the respective feature map using backpropagation. Our goal is to convert the model loss to pixel-wise uncertainty estimates. We assume that the pixel-wise uncertainty is in accordance with the depth estimation error. We later empirically show our claims in our evaluations.

In general, the most meaningful loss function is the error between the depth prediction $\mathbf{d}$ and the ground-truth depth $\mathbf{y} \in \mathbb{R}^{w \times h \times 1}$. Nevertheless, we have access to ground-truth information during training, but not during inference. Therefore, we define an auxiliary loss function $\mathcal{L}$ for the gradient generation. For the definition of this loss function, we consider the reference image $\mathbf{x}^{\prime}$ whose structure equals the structure of the original image $\mathbf{x}$. We make the assumption that the depth maps $\mathbf{d}$ and $\mathbf{d}^{\prime}$ predicted for $\mathbf{x}$ and $\mathbf{x}^{\prime}$ match. 
To generate the reference image $\mathbf{x}^{\prime}$, we define the horizontal flip operation $T(\cdot)$, which performs a left-right flip. 
The inverse function $T^{-1}(\cdot)$ reverts the flip operation. The depth prediction on an image $\mathbf{x}$ and the horizontally flipped version $\mathbf{x}^{\prime} = T(\mathbf{x})$ should be consistent, since the linear transformation preserves the pixel information of the original image and thus the structure of the scene. 
 
Overall, the reference depth $\mathbf{d}_{r}$ is the depth $\mathbf{d}^{\prime}$ predicted for the flipped image $\mathbf{x}^{\prime}$, which is flipped back to match the depth $\mathbf{d}$ obtained for the original input image $\mathbf{x}$. More precisely, we conduct two forward passes, one with the original image $\mathbf{x}$ to obtain $\mathbf{d} = f(\mathbf{x}; \theta)$ and one with the flipped version $\mathbf{x}^{\prime}=T(\mathbf{x})$ to obtain $\mathbf{d}^{\prime} = f(\mathbf{x}^{\prime}; \theta)$. Then, we apply the inverse function $T^{-1}(\mathbf{d}^{\prime})$ to obtain the reference depth $\mathbf{d}_{r}$. The auxiliary loss function is then defined based on the difference between the depth $\mathbf{d}$ and the reference depth $\mathbf{d}_{r}$ (see Sec.~\ref{sec:impl_methods}).
To obtain the gradients, we calculate the derivative of the loss \textit{w.r.t} the respective feature map $\mathbf{a}_{i}$: 
\begin{equation}
    \mathbf{g}_{i} = \frac{\partial \mathcal{L}}{\partial \mathbf{a}_{i}},
\end{equation}
where $\mathbf{g}_{i} \in \mathbb{R}^{w_{g} \times h_{g} \times c}$ with $w_{g}$, $h_{g}$ and $c$ being the feature map width, feature map height and the channel number, respectively. 

\paragraph{\textbf{Uncertainty Score}}
Next, the pixel-wise uncertainty estimate $\mathbf{u}$ is determined from the gradients $\mathbf{g}_{i}$ extracted from the neural network. Since an uncertainty value is required for each pixel, the gradient map must match the resolution of the depth map. 
First, we define the function $max_c(\cdot): \mathbb{R}^{w_{g} \times h_{g} \times c} \to \mathbb{R}^{w_{g} \times h_{g} \times 1}$ to perform max-pooling over the channel dimensions:  
\begin{equation}
    \mathbf{g}_{i}^{(\mathrm{max)}} = max_c(\mathbf{g}_{i}), 
\end{equation}
where $ \mathbf{g}_{i}^{(\mathrm{max)}} \in \mathbb{R}^{w_{g} \times h_{g} \times 1}$. We are interested in gathering the gradients with the largest magnitude for the uncertainty estimation using the max-pooling operation. Afterward, we define the function $upsample(\cdot): \mathbb{R}^{w_{g} \times h_{g} \times 1} \to \mathbb{R}^{w \times h \times 1}$ to upsample $\mathbf{g}_{i}^{(\mathrm{max})}$ by nearest-neighbor interpolation. Finally, we propose the uncertainty estimate as the self-normalized gradient map:
\begin{equation}
    \mathbf{u} = \frac{upsample(\mathbf{g}_{i}^{\mathrm{(max)}}) - \min\mathbf{g}_{i}^{\mathrm{(max)}}}{\max\mathbf{g}_{i}^{\mathrm{(max)}} - \min\mathbf{g}_{i}^{\mathrm{(max)}}},
\end{equation}
where $\max\mathbf{g}_{i}^{\mathrm{(max)}}$ and $\min\mathbf{g}_{i}^{\mathrm{(max)}}$ are the maximum and the minimum value of the gradient map, respectively. 
The final uncertainty is then normalized to a range of $[0, 1]$.

\subsection{Training Strategy}\label{sec:impl_methods}
We instantiate the auxiliary loss function for gradient generation based on the predicted depth $\mathbf{d}$ and the reference depth $\mathbf{d}_{r}$.
First, we define our loss function for depth estimation models that only predict the depth $\mathbf{d} = f_{r}(\mathbf{x}; \theta_{r})$. Then, we consider a Bayesian depth estimation model $\mathbf{d}, \mathbf{\sigma} = f_{b}(\mathbf{x}; \theta_{b})$ with $\mathbf{d}$ being the depth and $\mathbf{\sigma}$ the variance as a measure of uncertainty.  

\paragraph{\textbf{Depth Estimation Model}}
For standard depth estimation models $f_{\theta_r}$, that do not predict the uncertainty but only the depth, the auxiliary loss function is defined as follows: 
\begin{equation}
    \mathcal{L}_{r} = (\mathbf{d} - \mathbf{d}_{r})^{2}.
\end{equation}
An uncertainty estimate can be generated by a post-processing step in which image flipping is applied to estimate the uncertainty as the variance over two outputs (\textit{Post})~\cite{monodepth2}. Here, the variance is also the pixel-wise difference of the two depth predictions.
We claim that the information given by the gradients of the feature maps is higher compared to the uncertainty solely given by the loss. 
This is supported by the results in Sec.~\ref{sec:exp_results}, where we empirically demonstrate that the information given by the gradients of the feature maps is more effective for uncertainty estimation compared to simply relying on image flipping as a post-processing step. 

\paragraph{\textbf{Bayesian Depth Estimation Model}}
Since our approach is independent of the underlying training strategy, we do not only consider conventional depth estimation models but also the Bayesian version $f_{\theta_b}$ that already predicts the variance as an uncertainty measure.
Different strategies are proposed to train Bayesian models. One is training with log-likelihood maximization objective (\textit{Log}), while another is the self-teaching paradigm (\textit{Self}) as proposed by Poggi et al.~\cite{Poggi2020OnTU}. For Bayesian depth prediction networks, we aim to improve the uncertainty estimate given by the variance. Since these models do not only output the pixel-wise depth $\mathbf{d}$ but also the pixel-wise variance $\mathbf{\sigma}$, we also make use of the variance for the gradient generation. In this context, we add the squared variance to the loss term.
The auxiliary loss for the Bayesian model is given in the following: 
\begin{equation}
    \mathcal{L}_{b} = \mathcal{L}_{r} + \lambda \mathbf{\sigma}^{2},
\end{equation}
where $\lambda$ controls the influence of the variance loss term. Since $\mathbf{\sigma}$ represents the error inherent in the data, high variance leads to larger gradients, while small variance leads to small gradients. 

%% file: Sections/experiments.tex
\section{Experiments}
We evaluate our approach on two standard monocular depth estimation benchmarks. First, we describe the experimental setup including datasets, models, metrics, and implementation. Then, we demonstrate our uncertainty estimation results and provide visual illustrations.
Ultimately, we demonstrate the informativeness of the gradient space when ground truth is given for the loss computation and validate horizontal flipping as the chosen image transformation. 
\subsection{Experimental Setup}
\paragraph{\textbf{Datasets}}
Since there is no other work that performs a detailed analysis of different approaches to uncertainty estimation using a well-defined protocol, we follow the same evaluation as proposed by Poggi et al.~\cite{Poggi2020OnTU}. Therefore, we evaluate our approach on the  KITTI~\cite{Geiger2013IJRR} dataset. KITTI is an autonomous driving dataset taken at 61 scenes with an average image resolution of $375 \times 1242$. Similar to Poggi et al.~\cite{Poggi2020OnTU}, we use the Eigen split~\cite{Eigen2014DepthMP} with maximum depth set to 80 meters and use the improved ground truth depth provided in~\cite{Uhrig2017SparsityIC} for evaluation. In addition, we propose to evaluate the uncertainty estimation on the NYU Depth V2 dataset.  
This is an indoor depth estimation dataset taken at 494 scenes. The original image resolution is $480 \times 640$. Here, we adapt the evaluation protocol proposed by Poggi et al.~\cite{Poggi2020OnTU} but set the maximum depth to 10 meters.  
\paragraph{\textbf{Models}}
We choose Monodepth2~\cite{monodepth2} as the depth estimation model to make a fair comparison to the numerous uncertainty estimation approaches in~\cite{Poggi2020OnTU}. Similar to Poggi et al.~\cite{Poggi2020OnTU}, we use both Monodepth2 trained with monocular and stereo supervision.
Importantly, networks trained with monocular sequences are not only subject to depth uncertainty but also the uncertainty of the estimated camera pose. 
For NYU Depth V2~\cite{SilbermanECCV12} we train Monodepth2~\cite{monodepth2} in a supervised manner with the provided ground truth depth maps.
\paragraph{\textbf{Evaluation Metrics}}
Similar to prior work~\cite{Ilg2018UncertaintyEA,Poggi2020OnTU}, our goal is to estimate the uncertainty with respect to the prediction error. Therefore, we evaluate the estimated uncertainty on in-distribution data by \textit{sparsification} plots that measure if the uncertainty agrees with the true error.
Given a data sample and an error metric, the uncertainty is ranked in descending order. Then, the most uncertain pixels are removed to calculated the error metric for the remaining pixels to obtain the sparsification over the fraction of removed pixels. As in~\cite{Ilg2018UncertaintyEA,Poggi2020OnTU}, we compute the Area Under the Sparsification Error (AUSE) that is the difference between the sparsification and the oracle sparsification. The oracle sparsification is given if the uncertainty ranking corresponds to the ranking of the true error. We evaluate the AUSE in terms of the depth estimation metrics Absolute Relative Error (Abs Rel), Root Mean Squared Error (RMSE), and Accuracy ($\delta \geq 1.25$).  
In addition, as proposed by Poggi et al.~\cite{Poggi2020OnTU}, we compute the Area Under the Random Gain (AURG) that considers the difference between the sparsification and the random sparsification that quantifies whether the estimated uncertainty is better than not modeling the uncertainty. Again, we calculate the AURG in terms of Abs Rel, RMSE, and $\delta \geq 1.25$.
\paragraph{\textbf{Comparison to Related Work}}
Since our approach is a \textit{post hoc} uncertainty estimation method, we apply our approach to already trained depth estimation models. As described in Sec.~\ref{sec:impl_methods}, the base models (Base) are a standard depth estimation model with post-processing applied (\textit{Post})~\cite{monodepth2}, and the two Bayesian depth estimation models \textit{Log}~\cite{Poggi2020OnTU} and \textit{Self}~\cite{Poggi2020OnTU}, that also predict the variance.   
We compare our method to the uncertainties obtained from the base models themselves as well as inference only dropout (\textit{In-Drop})~\cite{Mi2019TrainingFreeUE}, MC Dropout (\textit{Drop})~\cite{mcdropout} and bootstrapped ensembles (\textit{Boot})~\cite{Lakshminarayanan2017SimpleAS}. Note that \textit{In-Drop} is also applied to the base models as a training-free approach, while \textit{Drop} and \textit{Boot} are separate models. Furthermore, we compare our approach to the uncertainty obtained with the variance over different test-time augmentations (\textit{Var}) as additional \textit{post hoc} baseline. 
\paragraph{\textbf{Implementation Details}}
All base models are implemented based on Monodepth2~\cite{monodepth2} as depth estimation network. Since Poggi et al.~\cite{Poggi2020OnTU} provide trained models for the uncertainly estimation baselines \textit{Post}, \textit{Log}, \textit{Self}, \textit{Boot}, and \textit{Drop} trained with monocular as well as stereo supervision, we evaluate our model with the provided weights for a fair comparison. For \textit{In-Drop} we apply dropout with probability $0.2$ at the same locations as used for \textit{Drop} but only during inference. As in~\cite{Poggi2020OnTU}, we perform $8$ forward passes for \textit{Drop} and \textit{In-Drop}.
For NYU Depth V2~\cite{SilbermanECCV12}, we train Monodepth2~\cite{monodepth2} as the base model with random rotation, random scaling, horizontal flipping, color jittering and cropped to a final resolution of $224 \times 288$. 
For this setup, we additionally implement the Bayesian model trained with log-likelihood maximization (\textit{Log}) as well as MC Dropout (\textit{Drop}) next to the standard depth estimation network. Moreover, \textit{In-Drop} is implemented as for KITTI~\cite{Geiger2013IJRR}. For \textit{Var}, we apply grayscale, flipping, noise and rotation as augmentations to calculate the variance over the resulting outputs. 
For all setups, we choose the 6th layer of the decoder to extract the gradients for our gradient-based uncertainty estimation approach. Results with gradients extracted from different layers are provided in the supplementary material.  
In the case of the Bayesian depth estimation networks, we choose  $\lambda$ to be $2.0$.
\subsection{Uncertainty Estimation Results}\label{sec:exp_results}
\begin{figure}[!ht]
    \centering
    \subfloat[KITTI Monocular Supervision\label{fig:kitti_mono}]{
    \centering
    \includegraphics[width=0.48\textwidth]{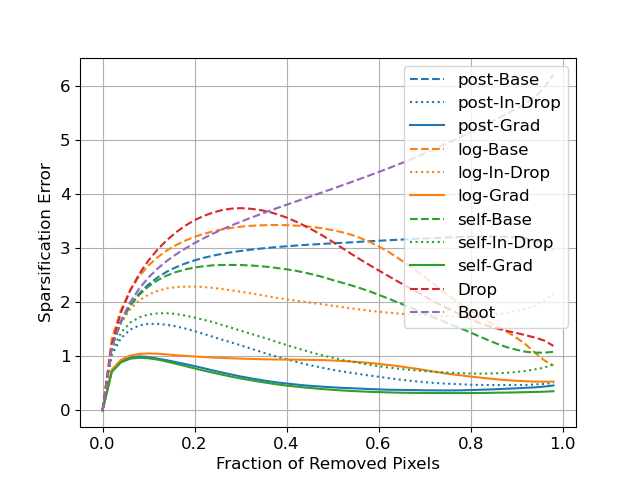}
    }
    \subfloat[KITTI Stereo Supervision\label{fig:kitti_stereo}]{
    \centering
    \includegraphics[width=0.48\textwidth]{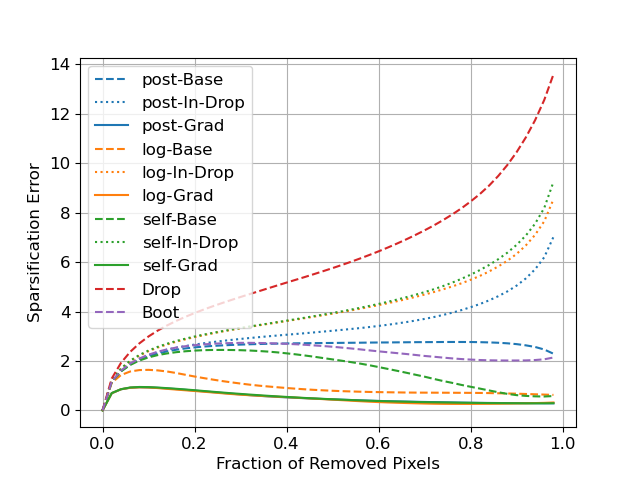}
    }\\
    \caption{The sparsification error in terms of RMSE over the fraction of removed pixels is shown for Monodepth2~\cite{monodepth2} trained on KITTI~\cite{Geiger2013IJRR} with monocular (a) and stereo (b) supervision.}
    \label{fig:spars_plots}
\end{figure}

\begin{figure}[ht]
\begin{tabular}{lll}
    \subfloat[RGB Image\label{fig:nyu_post_rgb}]{
    \includegraphics[width=0.28\textwidth]{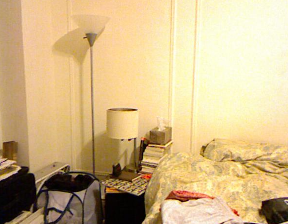}} & 
    \subfloat[Depth\label{fig:nyu_post_depth}]{
    \includegraphics[width=0.28\textwidth]{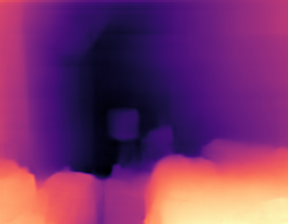}}
    \includegraphics[height=77pt]{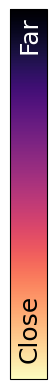} & 
    \subfloat[RMSE\label{fig:nyu_post_rmse}]{
    \includegraphics[width=0.28\textwidth]{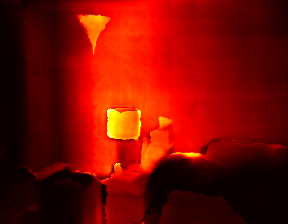}} 
    \includegraphics[height=77pt]{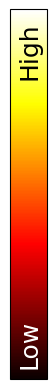} \\
     \subfloat[Post\label{fig:nyu_post_post}]{
    \includegraphics[width=0.28\textwidth]{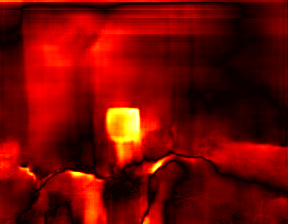}} &
    \subfloat[In-Drop\label{fig:nyu_post_drop}]{
    \includegraphics[width=0.28\textwidth]{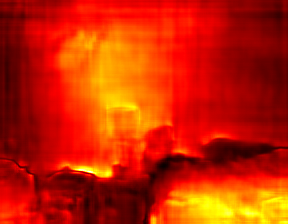}} & 
    \subfloat[Ours\label{fig:nyu_post_grad}]{
    \includegraphics[width=0.28\textwidth]{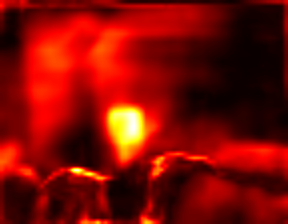}}
    \includegraphics[height=77pt]{Images/colorbar_hot.png} \\
\end{tabular}
    \caption{Uncertainty estimation example from Monodepth2~\cite{monodepth2} trained on NYU Depth V2~\cite{SilbermanECCV12}. In (a), the input image is shown. The depth prediction (b) and the root mean squared error (c) highlight that the lamp and the region around it are not estimated correctly. The second row demonstrates the uncertainty estimated by post-processing (d), inference dropout (e) and gradients (f).}
    \label{fig:nyu_post_example}
\end{figure}

\paragraph{\textbf{KITTI Monocular Supervision}}
The uncertainty estimation results for Monodepth2 trained with monocular sequences from KITTI are reported in Tab.~\ref{tab:kitti_mono}. For the monocular setup, where the uncertainty stems not only from the depth estimation model but also from the pose prediction, our method outperforms the base models as well as \textit{In-Drop}. Especially, AUSE and AURG in terms of RMSE are clearly improved compared to the base models \textit{Post}~\cite{monodepth2}, \textit{Log}~\cite{Poggi2020OnTU}, and \textit{Self}~\cite{Poggi2020OnTU}. Moreover, all methods obtain better uncertainty estimation results compared to the two empirical methods \textit{Drop}~\cite{mcdropout} and \textit{Boot}~\cite{Lakshminarayanan2017SimpleAS}. In terms of inference time, the base methods are slightly faster, but our gradient-based uncertainty estimation achieves better uncertainty estimation results in comparison. \textit{In-Drop}, on the other hand, has significantly higher inference times due to the sampling procedure and does not always outperform the baseline model when considering AUSE and AURG in terms of Abs Rel and $\delta \geq 1.25$. 
\setlength{\tabcolsep}{2pt}
\begin{table}[ht]
    \centering
    \caption{Uncertainty evaluation results for Monodepth2~\cite{monodepth2} trained with monocular supervision on KITTI~\cite{Geiger2013IJRR}.}
    \begin{tabular}{llccccccc}
        \hline
         \multicolumn{2}{c}{} & \multicolumn{2}{c}{Abs Rel} & \multicolumn{2}{c}{RMSE} & \multicolumn{2}{c}{$\delta \geq 1.25$} & \\
        \hline 
        Model & Method & AUSE $\downarrow$ & AURG $\uparrow$ &  AUSE $\downarrow$ & AURG $\uparrow$ &  AUSE $\downarrow$ & AURG $\uparrow$ & Inf [ms] \\
        \hline
        Drop & Base \cite{Poggi2020OnTU} & 0.056 & 0.000 & 2.568 & 0.944 & 0.097 & 0.002 & 47.12 \\
        Boot & Base \cite{Poggi2020OnTU} & 0.058 & 0.001 & 3.982 & -0.743 & 0.084 & -0.001 & 47.12 \\
        \hline
        \multirow{4}{*}{Post} & Base \cite{monodepth2} & 0.044 & 0.012 & 2.864 & 0.412 & 0.056 & 0.022 & 11.78 \\
         & Var & 0.055 & 0.003 & 3.575 & -0.210 & 0.074 & 0.006 & 29.45 \\
         & In-Drop \cite{Mi2019TrainingFreeUE} & 0.031 & 0.027 & 0.871 & 2.495 & 0.029 & 0.051 & 53.01 \\
         & Ours & \textbf{0.029} & \textbf{0.029} & \textbf{0.540} & \textbf{2.825} & \textbf{0.025} & \textbf{0.054} & 18.84 \\
        \hline
        \multirow{4}{*}{Log} & Base \cite{Poggi2020OnTU} & 0.039 & 0.020 & 2.562 & 0.916 & 0.044 & 0.038 & 5.89 \\
         & Var & 0.055 & 0.004 & 3.686 & -0.209 & 0.075 & 0.008 & 29.45 \\ 
         & In-Drop \cite{Mi2019TrainingFreeUE} & 0.045 & 0.014 & 1.916 & 1.561 & 0.056 & 0.027 & 53.01 \\
         & Ours & \textbf{0.026} & \textbf{0.033} & \textbf{0.819} & \textbf{2.658} & \textbf{0.024} & \textbf{0.059} & 18.84 \\
        \hline
        \multirow{4}{*}{Self} & Base \cite{Poggi2020OnTU} &  0.030 & 0.026 & 2.009 & 1.266 & 0.030 & 0.045 & 5.89 \\
         & Var & 0.055 & 0.001 & 3.632 & -0.357 & 0.073 & 0.002 & 29.45 \\
         & In-Drop \cite{Mi2019TrainingFreeUE} & 0.033 & 0.024 & 1.091 & 2.184 & 0.031 & 0.044 & 53.01 \\
         & Ours & \textbf{0.024} & \textbf{0.032} & \textbf{0.494} & \textbf{2.780} & \textbf{0.017} & \textbf{0.057} & 18.84 \\
        \hline 
    \end{tabular}
    \label{tab:kitti_mono}
\end{table}
\setlength{\tabcolsep}{2pt}
\begin{table}[ht]
    \centering
    \caption{Uncertainty evaluation results for Monodepth2~\cite{monodepth2} trained with stereo pair supervision on KITTI~\cite{Geiger2013IJRR}.} 
    \begin{tabular}{llccccccc}
        \hline
         \multicolumn{2}{c}{} & \multicolumn{2}{c}{Abs Rel} & \multicolumn{2}{c}{RMSE} & \multicolumn{2}{c}{$\delta \geq 1.25$} & \\
        \hline 
        Model & Method & AUSE $\downarrow$ & AURG $\uparrow$ &  AUSE $\downarrow$ & AURG $\uparrow$ &  AUSE $\downarrow$ & AURG $\uparrow$ & Inf [ms] \\
        \hline
        Drop & Base \cite{Poggi2020OnTU} & 0.103 & -0.029 & 6.163 & -2.169 & 0.231 & -0.080 & 47.12 \\
        Boot & Base \cite{Poggi2020OnTU} & 0.028 & 0.029 & 2.291 & 0.964 & 0.031 & 0.048 & 47.12 \\
        \hline
        \multirow{4}{*}{Post} & Base \cite{monodepth2} & 0.036 & 0.020 & 2.523 & 0.736 & 0.044  & 0.034 & 11.78 \\
         & Var & 0.054 & 0.003 & 3.672 & -0.327 & 0.074 & 0.006 & 29.45 \\
         & In-Drop \cite{Mi2019TrainingFreeUE} & 0.061 & -0.004 & 3.366 & -0.021 & 0.089 & -0.009 & 53.01 \\
         & Ours & \textbf{0.022} & \textbf{0.035} & \textbf{0.510} & \textbf{2.835} & \textbf{0.023} & \textbf{0.057} & 18.84 \\
        \hline
        \multirow{4}{*}{Log} & Base \cite{Poggi2020OnTU} & 0.022 & 0.036 & 0.938 & 2.402 & \textbf{0.018} & \textbf{0.061} & 5.89 \\
         & Var & 0.054 & 0.003 & 3.639 & -0.299 & 0.073 & 0.006 & 29.45 \\
         & In-Drop \cite{Mi2019TrainingFreeUE} & 0.073 & -0.016 & 4.063 & -0.724 & 0.107 & -0.028 & 53.01 \\
         & Ours & \textbf{0.019} & \textbf{0.038} & \textbf{0.490} & \textbf{2.849} & \textbf{0.018} & \textbf{0.061} & 18.84 \\
        \hline
        \multirow{4}{*}{Self} & Base \cite{Poggi2020OnTU} & \textbf{0.022} & \textbf{0.035} & 1.679 & 1.642 & \textbf{0.022} & \textbf{0.056} & 5.89 \\
         & Var & 0.056 & 0.001 & 3.728 & -0.408 & 0.076 & 0.002 & 29.45 \\
         & In-Drop \cite{Mi2019TrainingFreeUE} & 0.076 & -0.019 & 4.173 & -0.852 & 0.109 & -0.031 & 53.01 \\
         & Ours & \textbf{0.022} & \textbf{0.035} & \textbf{0.515} & \textbf{2.806} & 0.023 & 0.055 & 18.84 \\
        \hline 
    \end{tabular}
    \label{tab:kitti_stereo}
\end{table}
\paragraph{\textbf{KITTI Stereo Supervision}}
In Tab.~\ref{tab:kitti_stereo}, the uncertainty estimation results for Monodepth2 trained with stereo pair supervision, where the uncertainty results only from the depth estimation model, are listed. In this setup, the worst results are obtained from the computationally expensive MC Dropout~\cite{mcdropout}. For \textit{Post}~\cite{monodepth2} and \textit{Log}~\cite{Poggi2020OnTU}, the gradient-based uncertainty estimation especially improves over the base model in terms of AUSE and AURG for Abs Rel and RMSE. Moreover, our method is on par with the \textit{Self}~\cite{Poggi2020OnTU} model while being a \textit{post hoc} approach. One downside of self-teaching is the elaborate training procedure that is required to obtain the final depth estimation model. Furthermore, the training-free \textit{In-Drop}~\cite{Mi2019TrainingFreeUE} shows no improvement over the base models while requiring a significantly higher inference time. Furthermore, \textit{In-Drop}~\cite{Mi2019TrainingFreeUE} is barely better than random chance when observing AURG. 
\setlength{\tabcolsep}{2pt}
\begin{table}[ht]
    \centering
    \caption{Uncertainty evaluation results for Monodepth2~\cite{monodepth2} trained on NYU Depth V2~\cite{SilbermanECCV12}.} 
    \begin{tabular}{llccccccc}
        \hline 
         \multicolumn{2}{c}{} & \multicolumn{2}{c}{Abs Rel} & \multicolumn{2}{c}{RMSE} & \multicolumn{2}{c}{$\delta \geq 1.25$} & \\
        \hline 
        Model & Method & AUSE $\downarrow$ & AURG $\uparrow$ &  AUSE $\downarrow$ & AURG $\uparrow$ &  AUSE $\downarrow$ & AURG $\uparrow$ & Inf [ms] \\
        \hline
        Drop & Base \cite{mcdropout} & 0.086 & 0.002 & 0.214 & 0.160 & 0.169 & -0.001 & 26.32 \\
        \hline
        \multirow{4}{*}{Post} & Base \cite{monodepth2} & 0.066 & 0.018 & 0.267 & 0.096 & 0.116 & 0.037 & 6.58 \\
        & Var & 0.063 & 0.024 & 0.221 & 0.151 & 0.110 & 0.045 & 16.45 \\
         & In-Drop \cite{Mi2019TrainingFreeUE} & 0.086 & 0.000 & \textbf{0.223} & \textbf{0.149} & 0.163 & 0.000 &  29.61 \\
         & Ours & \textbf{0.061} & \textbf{0.025} & 0.252 & 0.120 & \textbf{0.106} & \textbf{0.048} & 12.05 \\
        \hline
        \multirow{4}{*}{Log} & Base \cite{Poggi2020OnTU} & 0.055 & 0.030 & \textbf{0.159} & \textbf{0.210} & 0.089 & 0.066 & 3.29 \\
         & Var & 0.061 & 0.024 & 0.220 & 0.149 & 0.106 & 0.049 & 16.45 \\
         & In-Drop \cite{Mi2019TrainingFreeUE} & 0.087 & -0.002 & 0.215 & 0.154 & 0.166 & -0.012 & 29.61 \\
         & Ours & \textbf{0.053} & \textbf{0.032} & 0.176 & 0.193 & \textbf{0.086} & \textbf{0.069} & 12.05 \\
        \hline 
    \end{tabular}
    \label{tab:nyu}
\end{table}
\paragraph{\textbf{NYU Depth V2}}
The uncertainty estimation performance of Monodepth2 trained on NYU Depth V2 is reported in Tab.~\ref{tab:nyu}. Again, the uncertainty only stems from the depth estimation itself due to the supervised training. We implement our approach on the \textit{Post}~\cite{monodepth2} and \textit{Log}~\cite{Poggi2020OnTU} models and state the MC Dropout~\cite{mcdropout} model as baseline. As in the previous setups, uncertainty estimation with dropout sampling results in the worst uncertainty estimation performance. Here, both versions, with dropout during training as well as only during inference, do obtain worse uncertainty estimates compared to the base models in almost all metrics. By contrast, our gradient-based method does improve the performance for AUSE and ARUG in terms of Abs Rel as well as $\delta \geq 1.25$. 
\setlength{\tabcolsep}{4pt}
\begin{table}[ht]
        \caption{Comparison of the different uncertainty estimation methods regarding number of trained models (\#Train), whether a specific training strategy is required (Specialized Training), number of models for uncertainty estimation (\#Models), number of forward passed required to obtain depth and uncertainty (\#Forward) and number of backward passes (\#Backward). Note that $N$ is set to $8$.}
    \centering
    \begin{tabular}{lccccc}
        \hline
        Method &  \#Train & Specialized Training & \#Models & \#Forward & \#Backward \\
        \hline
        Drop \cite{mcdropout} & 1 & Yes & 1 & N & - \\
        Boot \cite{Lakshminarayanan2017SimpleAS} & N & Yes & N & N & - \\
        \hline 
        Post \cite{monodepth2} & 1 & No & 1 & 2 & - \\
        Log \cite{Poggi2020OnTU} & 1 & Yes & 1 & 1 & - \\
        Self \cite{Poggi2020OnTU} & 2 & Yes & 1 & 1 & - \\ 
        \hline 
        Var & - & No & - & 4 + 1 & - \\
        In-Drop \cite{Mi2019TrainingFreeUE} & - & No & - & N + 1 & - \\
        Ours & - & No & - &  2 & 1 \\
        \hline
    \end{tabular}
    \label{tab:method_cmp}
\end{table}
\paragraph{\textbf{Sparsification Error Plots}}
In Fig.~\ref{fig:spars_plots}, we illustrate the sparsification error curves in terms of RMSE for KITTI with monocular supervision (Fig.~\ref{fig:kitti_mono}) and with stereo pair supervision (Fig.~\ref{fig:kitti_stereo}) averaged over the KITTI test set. The sparsification error represents the deviation of the sparsification curve from the respective oracle sparsification. Therefore, a smaller area under the curve means better performance in uncertainty estimation. For our \textit{post hoc} approach, the uncertainty best matches the true error, while \textit{In-Drop}~\cite{Mi2019TrainingFreeUE} does not improve on the baseline model for the stereo setup. In the monocular setup, \textit{Boot}~\cite{Lakshminarayanan2017SimpleAS} performs the worst, while the uncertainty modeled with \textit{Drop}~\cite{mcdropout} in the stereo case is the least consistent with the true error.  
\paragraph{\textbf{Summary}} 
Overall, our method improves the uncertainty estimates for most of the cases over the base model while only adding a slight overhead in inference time. The empirical approaches \textit{Boot}~\cite{Lakshminarayanan2017SimpleAS}, \textit{Drop}~\cite{mcdropout}, and \textit{In-Drop}~\cite{Mi2019TrainingFreeUE}, on the other hand, add a huge computational overhead and therefore require inference times that are not feasible in real-time applications such as autonomous driving or robotics. In addition, these methods do not outperform the uncertainty of the \textit{Post}~\cite{monodepth2}, \textit{Log}~\cite{Poggi2020OnTU}, and \textit{Self}~\cite{Poggi2020OnTU} models. The comparison of our method with the \textit{Post} baseline highlights the higher significance of gradients extracted from the model compared to using image flipping alone as the difference in pixel space. In Tab.~\ref{tab:method_cmp}, the main characteristics of the different methods are compared regarding the number of trained models (\#Train), the adaptation of the training strategy (Specialized Training), the number of models that must be saved (\#Models), the number of forward passes necessary to obtain the depth as well as the uncertainty estimate (\#Forward) and the number of required backward passes (\# Backward). In general, only \textit{In-Drop}, \textit{Var} and our method are \textit{post hoc} approaches applicable to all kinds of depth estimation networks. Moreover, all approaches besides post-processing require an adaptation to the training strategy or the network architecture. While the empirical methods result in a large computational overhead, our approach only needs one additional backward pass to obtain the gradients. Moreover, empirical methods are known to decrease the depth estimation performance \cite{Poggi2020OnTU}. When comparing our approach with the other training-free methods \textit{In-Drop}~\cite{Mi2019TrainingFreeUE} and \textit{Var}, these approaches need one forward pass to obtain the depth prediction and $N$ or $4$ (depending on the number of augmentations) forward passes to get the uncertainty estimate, while our approach only needs two forward and one backward pass to obtain depth and uncertainty estimates. 
\paragraph{\textbf{Visual Results}}
In Fig.~\ref{fig:nyu_post_example}, an example from Monodepth2 trained with the standard depth estimation protocol on NYU Depth V2 is illustrated. Overall, the estimated uncertainty for high error regions should be high, while the uncertainty for low error regions should be low. When comparing the depth prediction (Fig.~\ref{fig:nyu_post_depth}) and the RMSE (Fig.\ref{fig:nyu_post_rmse}), it becomes apparent that the region around the two lamps has the greatest error. Both, the post-processing~\cite{monodepth2} (Fig.~\ref{fig:nyu_post_post}) and our gradient-based method (Fig.~\ref{fig:nyu_post_grad}) highlight the high uncertainty at the lamp. Moreover, our approach improves over the uncertainty estimation by post-processing in regions with low error. \textit{In-Drop}~\cite{Mi2019TrainingFreeUE} (Fig.~\ref{fig:nyu_post_drop}), on the other hand, overestimates uncertainty in most regions, especially in the lower right corner. More visual results can be found in the supplementary material. 
\subsection{Ablation Study}
In Tab.~\ref{tab:ablation}, the uncertainty estimation results of our gradient-based approach on Monodepth2 trained with NYU Depth V2 are reported for the \textit{Post} model. We demonstrate different configurations to define the loss for the gradient generation. We consider the squared difference of the prediction to the ground truth depth (GT) and to transformed images by image flipping (Flip), gray-scale conversion (Gray), additive Gaussian noise (Noise) and rotation (Rot) with different rotation angles. Note that the selected augmentations are not necessarily within the training time augmentation. 
The uncertainty estimation results for all three depth estimation metrics clearly show that the loss between the prediction and the ground truth depth results in the best uncertainty estimation performance. Nevertheless, the ground truth is not available during inference in real-world application. However, results in Sec.~\ref{sec:exp_results} demonstrate promising results for uncertainty estimation based on the information given by the extracted gradients. Comparing the different transformation operations, the flipping operation performs best. 
Compared to the other augmentations considered, when flipping horizontally, the model observes the scene from a different context, while preserving the geometry of the scene. In case of the rotation operation, the uncertainty cannot be determined for each pixel. This confirms the use of the flipping operation to create a reference depth.
\setlength{\tabcolsep}{3pt}
\begin{table}[!ht]
    \centering
   \caption{Uncertainty estimation results for Monodepth2~\cite{monodepth2} Post trained on NYU Depth V2~\cite{SilbermanECCV12} when using different loss functions for the gradient generation. We compare the error of the prediction to the ground truth depth (GT) and depth predictions obtained by different image transformations. We consider image flipping (Flip), gray-scale conversion (Gray), additive Gaussian noise (Noise) and rotation (Rot-$[$angle$]$) with different rotation angles.}
    \begin{tabular}{llcccccc}
        \hline 
          & \multicolumn{2}{c}{Abs Rel} & \multicolumn{2}{c}{RMSE} & \multicolumn{2}{c}{$\delta \geq 1.25$} \\
        \hline 
        Loss & AUSE $\downarrow$ & AURG $\uparrow$ &  AUSE $\downarrow$ & AURG $\uparrow$ &  AUSE $\downarrow$ & AURG $\uparrow$ \\
        \hline
        GT & \textbf{0.020} & \textbf{0.066} & \textbf{0.106} & \textbf{0.266} & \textbf{0.031} & \textbf{0.123} \\
        Flip & \underline{0.061} & \underline{0.025} & \underline{0.252} & \underline{0.120} & \underline{0.106} & \underline{0.048} \\
        Gray & 0.067 & 0.020 & 0.253 & 0.119 & 0.111 & 0.044 \\
        Noise & 0.066 & 0.020 & 0.261 & 0.111 & 0.108 & 0.046 \\ 
        Rot-5° & 0.066 & 0.020 & 0.267 & 0.105 & 0.114 & 0.041 \\
        Rot-10° & 0.069 & 0.017 & 0.269 & 0.103 & 0.124 & 0.031 \\
        Rot-20° & 0.076 & 0.011 & 0.270 & 0.102 & 0.143 & 0.012 \\
        \hline
    \end{tabular}
    \label{tab:ablation}
\end{table}